\documentclass[11pt]{article}
\usepackage{acl2016}
\usepackage{times}
\usepackage{latexsym}
\usepackage{paralist}
\usepackage{url}

\usepackage{color}
\usepackage{graphicx}

\newcounter{tbsnr}
\newenvironment{tbs}
{\addtocounter{tbsnr}{1}\par\bigskip\noindent\fbox{\thetbsnr}
\hspace*{\fill}\begin{minipage}{7cm}\tt}
{\end{minipage}\hspace*{\fill}\bigskip}

\newcommand{\cut}[1]{}

\usepackage{array}
\newcolumntype{C}[1]{>{\centering\let\newline\\\arraybackslash\hspace{0pt}}m{#1}}

\aclfinalcopy 
\def\aclpaperid{327} 

\title{``The red one!'': \\On learning to refer to things \\based on discriminative properties
}

\author{Angeliki Lazaridou \and Nghia The Pham \and Marco Baroni\\
	    University of Trento\\
	    {\tt \{{angeliki.lazaridou|thepham.nghia|marco.baroni\}}@unitn.it}}

\date{}

\begin{document}

\maketitle

\begin{abstract}

  As a first step towards agents learning to communicate about their
  visual environment, we propose a system that, given visual
  representations of a referent (\url{CAT}) and a context
  (\url{SOFA}), identifies their \emph{discriminative attributes},
  i.e., properties that distinguish them (\url{has_tail}). Moreover,
  although supervision is only provided in terms of discriminativeness
  of attributes for pairs, the model learns to assign plausible
  attributes to specific objects (\url{SOFA}-\url{has_cushion}). Finally, we
  present a preliminary experiment confirming the referential success
  of the predicted discriminative attributes.
\end{abstract}

\section{Introduction}
\label{intro}

There has recently been renewed interest in developing systems capable
of genuine language understanding
\cite{Hermann:etal:2015,Hill:etal:2015}. In this perspective, it is
important to think of an appropriate general framework for teaching
language to machines. 
Since we use language primarily for communication,  a reasonable
approach is to develop systems within a genuine communicative setup
\cite{Steels:2003,Mikolov:etal:2015}.  Out long-term goal is thus to develop
communities of computational agents that learn how to use language efficiently
in order to achieve communicative success~\cite{Vogel:etal:2013,Foerster:etal:2016}.

\begin{figure}[t]
\center
\includegraphics[scale=0.25]{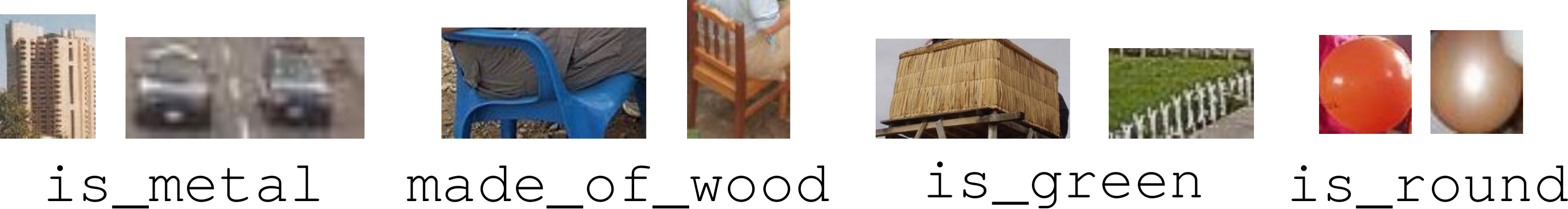}
\caption{Discriminative attributes predicted by our model. Can you
  identify the intended referent? See Section \ref{sec:experiment3}
  for more information}
\label{fig:quiz}
\end{figure}

Within this general picture, one fundamental aspect of meaning where
communication is indeed crucial is the act of
\emph{reference}~\cite{Searle:1969,Abbott:2010}, the ability to
successfully talk to others about things in the external world. A
specific instantiation of reference studied in this paper is that of
referring expression
generation~\cite{Dale:Reiter:1995,Mitchell:etal:2010,Kazemzadeh:etal:2014}. A
\emph{necessary} condition for achieving successful reference is that
referring expressions (REs) accurately distinguish the intended
referent from any other object in the
context~\cite{Dale:Haddock:1991}.    Along these lines, we present here
a model that, given an intended referent and a context object, 
predicts the attributes that discriminate between the two. Some examples
of the behaviour of the model are presented in Figure~\ref{fig:quiz}.

Importantly, and distinguishing our work from earlier
literature on generating REs~\cite{Krahmer:VamDeemter:2012}: 
\begin{inparaenum}[(i)]
\item the input objects are represented by natural images, so that the
  agent must learn to extract relevant attributes from realistic data;
  and
\item no direct supervision on the attributes of a single object is
  provided: the training signal concerns their discriminativeness for
   object pairs (that is, during learning, the agent might be
  told that \url{has_tail} is discriminative for $\langle$\url{CAT},
  \url{SOFA}$\rangle$, but not that it is an attribute of cats). %
  We use this ``pragmatic'' signal since it could later be replaced by
  a measure of success in actual communication between two agents
  (e.g., whether a second agent was able to pick the correct referent
  given a RE).
\end{inparaenum}



 

\section{Discriminative Attribute Dataset}
\label{sec:dataset}

We generated the Discriminative Attribute Dataset, consisting of
pairs of (intended) \emph{referents} and \emph{contexts}, with respect
to which the referents should be identified by their distinctive
attributes.

\begin{table}[t]
\small
\begin{center}
\begin{tabular}{ @{}l l C{3cm}}
\textbf{$\langle$referent,}  & \textbf{visual instances} & \textbf{discriminative}\\
~~~~\textbf{context$\rangle$}  &                         & \textbf{attributes}              \\\hline
$\langle$\url{CAT}, \url{SOFA} $\rangle$ & \includegraphics[scale=0.16]{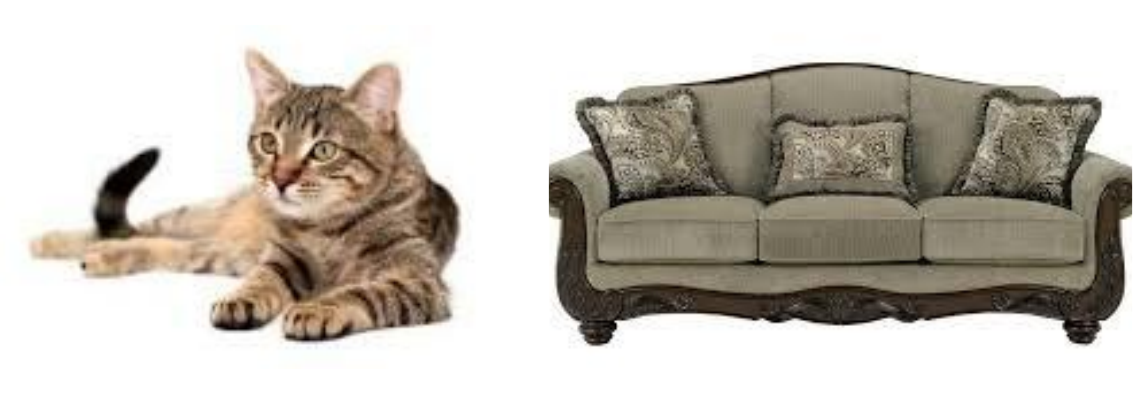} & has\_tail, has\_cushion, ...\\
$\langle$\url{CAT}, \url{APPLE}$\rangle$ &\includegraphics[scale=0.15]{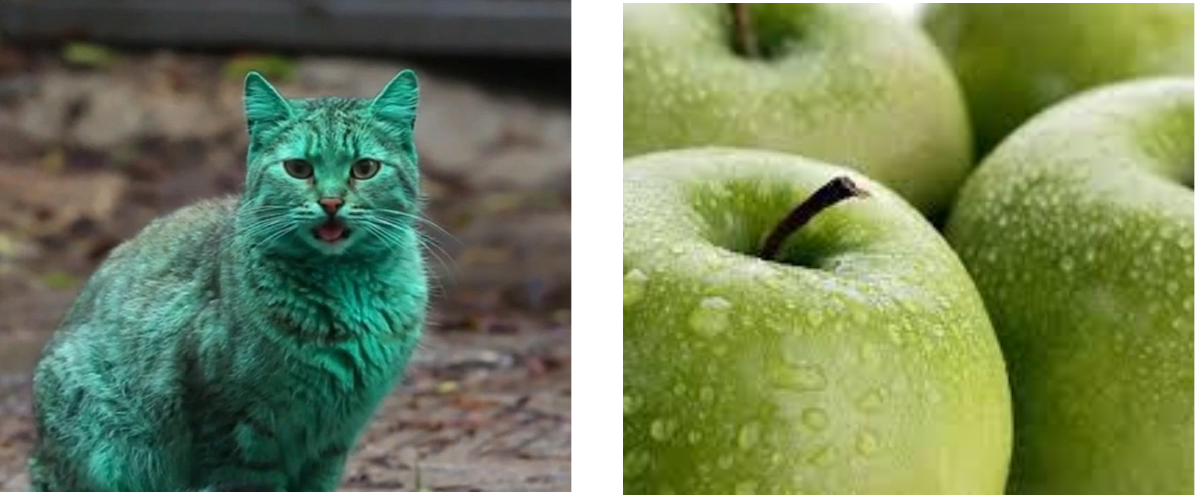}  & has\_legs, \textit{is\_green}, ...\\

\end{tabular}
\caption{Example training data}
\label{tab:dataset}
\end{center}
\end{table}

Our starting point is the Visual Attributes for Concepts Dataset
(ViSA)~\cite{Silberer:etal:2013}, which contains \emph{per-concept}
(as opposed to \emph{per-image}) attributes for 500 concrete concepts
(\url{CAT}, \url{SOFA}, \url{MILK}) spanning across different
categories (\url{MAMMALS}, \url{FURNITURE}), annotated with 636
general attributes.  We disregarded ambiguous concepts (e.g.,
\emph{bat}), thus reducing our working set of concepts $C$ to 462 and
the number of attributes $V$ to 573, as we eliminated any attribute
that did not occur with concepts in $C$.  We extracted on average 100
images annotated with each of these concepts from
ImageNet~\cite{Deng:etal:2009}. Finally, each image $i$ of concept $c$
was associated to a \emph{visual instance vector}, by feeding the
image to the VGG-19 ConvNet~\cite{Simonyan:Zisserman:2014}, as
implemented in the MatConvNet toolkit~\cite{Vedaldi:Lenc:2015}, and
extracting the second-to-last fully-connected (fc) layer as its
4096-dimensional visual representation $\mathbf{v}_{{c}_i}$.

We split the target concepts into \emph{training}, \emph{validation}
and \emph{test} sets, containing 80\%, 10\% and 10\% of the concepts
in each category, respectively. This ensures that
\begin{inparaenum}[(i)] \item the intersection between train and test
  concepts is empty, thus allowing us to test the generalization of
  the model across different objects, but \item there are instances of all
  categories in each set, so that it is possible to generalize across
  training and testing objects.
\end{inparaenum} Finally we build all possible combinations of
concepts in the training split to form pairs of referents and contexts
$\langle c_r, c_{c}\rangle$ and obtain their (binary) attribute
vectors $\mathbf{p}_{c_r}$ and $\mathbf{p}_{c_{c}}$ from ViSA,
resulting in 70K training pairs. From the latter, we derive, for each
pair, a concept-level ``discriminativeness'' vector by computing the
symmetric difference
$\mathbf{d}_{c_r,c_{c}}=(\mathbf{p}_{c_r}-\mathbf{p}_{c_{c}})\cup
(\mathbf{p}_{c_{c}}-\mathbf{p}_{c_r})$. The latter will contain 1s for
discriminative attributes, 0s elsewhere. On average, each pair is
associated with 20 discriminative attributes. The final training data
are triples of the form $\langle c_r, c_{c},
\mathbf{d}_{c_r,c_{c}}\rangle$ (the model \emph{never} observes the
attribute vectors of specific concepts), to be associated with visual
instances of the two concepts.  Table~\ref{tab:dataset} presents some
examples.

Note that ViSA contain concept-level attributes, but images contain
specific instances of concepts for which a general attribute might not
hold. This introduces a small amount of noise. For example,
\url{is_green} would in general be a discriminative attribute for
apples and cats, but it is not for the second sample in Table
\ref{tab:dataset}. Using datasets with \emph{per-image} attribute
annotations would solve this issue.  However, those currently available 
only cover specific classes of concepts (e.g., only clothes, or
animals, or scenes, etc.).  Thus, taken separately, they are not
general enough for our purposes, and we cannot merge them, since their
concepts live in different attribute spaces.

\section{Discriminative Attribute Network}

The proposed \emph{Discriminative Attribute Network} (DAN) learns to
predict the discriminative attributes of referent object $c_r$ and
context $c_{c}$ without direct supervision at the attribute level, but
relying only on discriminativeness information (e.g., for the objects
in the first row of Table \ref{tab:dataset}, the gold vector would
contain 1 for \url{has_tail}, but 0 for both \url{is_green} and
\url{has_legs}). Still, the model is implicitly encouraged to embed
objects into a consistent attribute space, to generalize across the
discriminativeness vectors of different training pairs, so it also
effectively learns to annotate objects with visual attributes.

\begin{figure}[t]
\center
\includegraphics[scale=0.60]{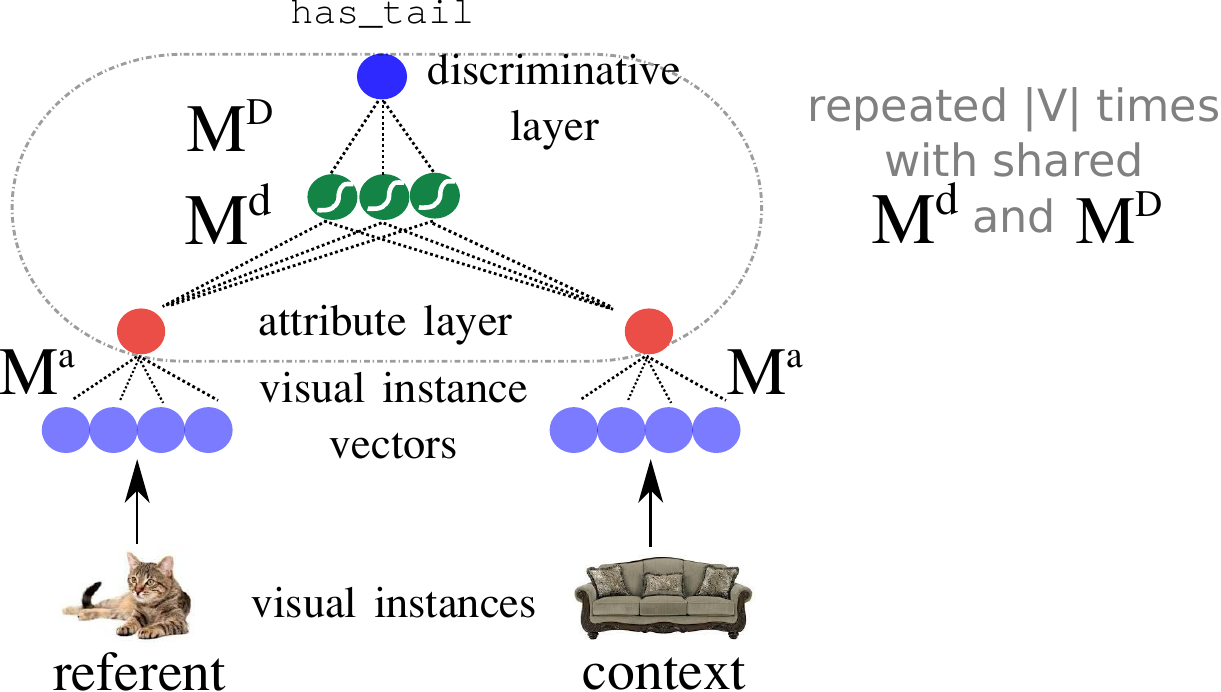}
\caption{Schematic representation of DAN. For simplicity, 
the prediction process is only illustrated for \url{has_tail}}
\label{fig:model}
\end{figure}

Figure~\ref{fig:model} presents a schematic view of DAN, focusing on a
single attribute. The model is presented with two concepts
$\langle$\url{CAT}, \url{SOFA}$\rangle$, and randomly samples a visual
instance of each.  The instance visual vectors $\mathbf{v}$ (i.e.,
ConvNet second-to-last fc layers) are mapped into \emph{attribute} vectors of
dimensionality $|V|$ (cardinality of all available attributes), using
weights $\mathbf{M}^\mathbf{a}\in\mathbf{R}^{4096\times{}|V|}$ shared
between the two concepts. Intuitively, this layer should learn
whether an attribute is active for a specific object, as this is
crucial for determining whether the attribute is discriminative for an
object pair. In Section~\ref{sec:exp2}, we present experimental
evidence corroborating this hypothesis.

In order to capture the \emph{pairwise} interactions between 
attribute vectors, the model proceeds by concatenating the two units
associated with the \emph{same} visual attribute $v$ across the two
objects (e.g., the units encoding information about \url{has_tail})
and pass them as input to the \emph{discriminative} layer.  The
discriminative layer processes the two units by applying a linear
transformation with weights
$\mathbf{M}^\mathbf{d}\in\mathbf{R}^{2\times{}h}$, followed by a
sigmoid activation function, finally deriving a single value by
another linear transformation with weights
$\mathbf{M}^\mathbf{D}\in\mathbf{R}^{h\times{}1}$.
The output $\hat{d}_v$ encodes the predicted degree of
discriminativeness of attribute $v$ for the specific reference-context
pair.  The same process is applied to all attributes $v\in V$, to
derive the estimated discriminativeness vector $\mathbf{\hat{d}}$,
using the same shared weights $\mathbf{M}^\mathbf{d}$ and
$\mathbf{M}^\mathbf{D}$ for each attribute.

To learn the parameters $\theta$ of the model (i.e. $\mathbf{M}^\mathbf{a}$,
$\mathbf{M}^\mathbf{d}$ and  $\mathbf{M}^\mathbf{D}$), given training data 
 $\langle c_r, c_{c}, \mathbf{d}_{c_r,c_{c}}\rangle$, we
 minimize MSE between the gold vector $\mathbf{d}_{c_r,c_{c}}$ and model-estimated
$\mathbf{\hat{d}}_{c_r,c_{c}}$. 
We trained the model with rmsprop and with a batch size of 32.
All hyperparameters (including the hidden size $h$ which was set to 60)  were
tuned to maximize performance on the validation set.


\section{Predicting Discriminativeness}
\label{sec:exp1}

We evaluate the ability of the model to
predict attributes that discriminate the intended referent from the
context. Precisely, we ask the model to return all
\emph{discriminative attributes} for a pair, independently of whether
they are positive for the referent or for the context (given images of
a cat and a building, both $+$\url{is_furry} and $-$\url{made_of_bricks}
are discriminative of the cat).

\paragraph{Test stimuli} We derive our test stimuli from the VisA test
split (see Section~\ref{sec:dataset}), containing 2000 pairs.
Unlike in training, where the model was presented with
specific \emph{visual instances} (i.e., single images), for evaluation
we use \emph{visual concepts} (\url{CAT}, \url{BED}), which we derive
by \emph{averaging} the vectors of all images associated to an object
(i.e., deriving \url{CAT} from all images of cats), due to lack of
gold information on \emph{per-image} attributes.  

\paragraph{Results} We compare DAN against a random baseline based on
per-attribute discriminativeness probabilities estimated from the
training data and an ablation model without attribute layer. We test
moreover a model that is trained with supervision to predict
attributes and then deterministically computes the discriminative
attributes. Specifically, we implemented a neural network with one
hidden layer, which takes as input a visual instance, and it is
trained to predict its gold attribute vector, casting the problem as
logistic regression, thus relying on supervision at the attribute
level. Then, given two paired images, we let the model generate their
predicted attribute vectors and compute the discriminative attributes
by taking the symmetric difference of the predicted attribute vectors
as we do for DAN.  For the DAN and its ablation, we use a 0.5
threshold to deem an attribute discriminative, without
tuning.  

The results in Table~\ref{tab:exp1} confirm that, with appropriate supervision,
DAN performs discriminativeness prediction reasonably well -- indeed, as well as the model 
with similar parameter capacity requiring direct supervision on an attribute-by-attribute basis, followed by the symmetric difference calculation. 
Interestingly,
allowing the model to embed visual representations into an intermediate
attribute space has a strong positive effect on performance. Intuitively, since
DAN is evaluated on novel concepts, the mediating attribute layer provides more
high-level semantic information helping generalization, at the expense
of extra parameters compared to the ablation without attribute layer.



\begin{table}
\small
\begin{center}
\begin{tabular}{ c||c|c|c } 
Model & Precision & Recall&F1 \\\hline
DAN  & 0.66 & 0.49 & 0.56 \\ 
attribute+sym.~difference & 0.64 & 0.48 & 0.55 \\
no attribute layer & 0.63& 0.33 & 0.43\\\hline 
Random baseline &0.16 &0.16 &0.16 \\
\end{tabular}
\caption{Predicting discriminative features}
\label{tab:exp1}
\end{center}
\end{table}

\section{Predicting Attributes}
\label{sec:exp2}
  
Attribute learning is typically studied in supervised
setups~\cite{Ferrari:Zisserman:2007,Fahradi:etal:2009,Russakovsky:Fei-Fei:2010}.
Our model learns to embed visual objects in an attribute space through
indirect supervision about attribute discriminativeness for specific
$<$\emph{referent}, \emph{context}$>$ pairs. Attributes are
\emph{never} explicitly associated to a specific concept during
training. The question arises of whether discriminativeness pushes the
model to learn plausible concept
attributes. 
Note that the idea that the semantics of attributes arises from their
distinctive function within a communication system is fully in line
with the classic structuralist view of linguistic meaning
\cite{Geeraerts:2009}.

To test our hypothesis, we feed DAN the same test stimuli (visual
concept vectors) as in the previous experiment, but now look at
activations in the attribute layer.  Since these activations are real
numbers whereas gold values (i.e., the visual attributes in the ViSA
dataset) are binary, we use the validation set to learn the threshold
to deem an attribute active, and set it to 0.5 without tuning.  Note
that no further training and no extra supervision other than the
discriminativeness signal are needed to perform attribute prediction.
The resulting binary attribute vector $\mathbf{\hat{p}}_c$ for concept
$c$ is compared against the corresponding gold attribute vector
$\mathbf{p}_c$.

\paragraph{Results} We compare DAN to the random baseline and to an explicit \emph{attribute classifier} similar
to the one used in the previous experiment, i.e., a one-hidden-layer neural network trained with logistic regression to predict 
the attributes.
We report moreover the best F1 score of ~\newcite{Silberer:etal:2013},
who learn a SVM for each visual attribute based on HOG visual features. Unlike
in our setup, in theirs, images for the same concept are used  both for training and to derive visual attributes (our setup is ``zero-shot'' at the concept level,
i.e., we predict attributes of concepts not seen in training).  Thus, despite the fact that they used presumably less accurate pre-CNN visual features, the setup is much easier for them, and we take their performance to be an upper bound on ours.

DAN reaches, and indeed surpasses, the performance of the
model with direct supervision at the attribute level, confirming the power of
discriminativeness as a driving force in building semantic representations. The
comparison with Silberer's model suggests that there is room
for improvement, although the noise inherent in concept-level annotation
imposes a relatively low bound on realistic performance.
\begin{table}
\small
\begin{center}
\begin{tabular}{ c||c|c|c }
Model & Precision & Recall&F1 \\\hline
DAN & 0.58 & 0.64&0.61\\
direct supervision & 0.56& 0.60 & 0.58\\\hline
Silberer et. al.&0.70 &0.70 &0.70\\
Random baseline & 0.13 & 0.12&  0.12\\
\end{tabular}
\caption{Predicting concept attributes}
\end{center}
\end{table}

\section{Evaluating Referential Success}
\label{sec:experiment3}

We finally ran a pilot study testing whether DAN's ability to predict
discriminative attributes at the concept level translates into
producing features that would be useful in constructing successful
referential expressions for specific object instances.

\paragraph{Test stimuli} Our starting point is the ReferIt
dataset~\cite{Kazemzadeh:etal:2014}, consisting of REs denoting objects (delimited by bounding boxes) in
natural images.  We filter out any $\langle$RE, bounding box$\rangle$
pair whose RE does not overlap with our attribute set $V$ and annotate
the remaining ones with the overlapping attribute, deriving data of
the form $\langle$RE, bounding box, \url{attribute}$\rangle$.  For
each intended referent of this type, we sample as context another
$\langle$RE, bounding box$\rangle$ pair such
that \begin{inparaenum}[(i)] 
\item the context RE does not contain the referent \url{attribute}, so
  that the latter is a likely discriminative feature;
\item referent and context come from different images, so that their
  bounding boxes do not accidentally overlap;
\item there is maximum word overlap between
  referent and contexts REs, creating a realistic referential
  ambiguity setup (e.g., two cars, two objects in similar
  environments).
\end{inparaenum} 
Finally we sample maximally 20 $\langle$referent, context$\rangle$
pairs per \url{attribute}, resulting in 790 test items.  For each
referent and context we extract CNN visual vectors from their bounding
boxes, and feed them to DAN to obtain their discriminative
attributes. Note that we used the ViSA-trained DAN for this experiment as
well.

\paragraph{Results} For 12\% of the test $\langle$referent,
context$\rangle$ pairs, the discriminative \url{attribute} is
contained in the set of discriminative attributes predicted by DAN. A
random baseline estimated from the distribution of attributes in the
ViSA dataset would score 15\% recall. This baseline however
on average predicts 20 discriminative attributes, 
whereas DAN activates, only 4. Thus, the baseline has a trivial recall advantage.

In order to evaluate whether in general the discriminative attributes
activated by DAN would lead to referential success, we further sampled
a subset of 100 $\langle$referent, context$\rangle$ test pairs. We
presented them separately to two subjects (one a co-author of this
study) together with the attribute that the model activated with the
largest score (see Figure~\ref{fig:quiz} for examples). Subjects were
asked to identify the intended referent based on the attribute. If
both agreed on the same referent, we achieved referential success,
since the model-predicted attribute sufficed to coherently
discriminate between the two images. 
Encouragingly, the subjects agreed on 78\% of the pairs (p$<$0.001
when comparing against chance guessing, according to a 2-tailed
binomial test). In cases of disagreement, the predicted attribute was
either too generic or very salient in both objects, a behaviour
observed especially in same-category pairs (e.g., \url{is_round} in
Figure~\ref{fig:quiz}).

\section{Concusion}
\label{sec:conclusion}
We presented DAN, a model that, given a referent and a context, learns
to predict their discriminative features, while also inferring visual
attributes of concepts as a by-product of its training regime.  While
the predicted discriminative attributes can result in referential
success, DAN is currently lacking all other properties of reference~\cite{Grice:1975}
(salience, linguistic and pragmatic felicity, etc).  We are currently working
towards adding communication (thus simulating a speaker-listener
scenario~\cite{Golland:etal:2010}) and natural language to the picture.

\section*{Acknowledgments}

This work was supported by
ERC 2011 Starting Independent Research Grant
n.~283554 (COMPOSES). We gratefully acknowledge the
support of NVIDIA Corporation with the donation of the
GPUs used for this research.

\bibliography{../../angeliki.bib,../../marco.bib}
\bibliographystyle{acl2016}

\end{document}